\documentclass[11pt,a4paper]{article}
\usepackage[hyperref]{acl2018}
\usepackage{times}
\usepackage{latexsym}

\usepackage{url}
\usepackage{setspace}
\usepackage{graphicx}
\usepackage{amsmath}
\usepackage{amsfonts}
\usepackage{algorithm}
\usepackage[noend]{algorithmic}
\usepackage{multirow}
\usepackage{array}

\aclfinalcopy 

\title{Learning to Generate Structured Queries from Natural Language with Indirect Supervision}

\name{Ziwei Bai$^1$, Bo Yu$^2$, Bowen Wu$^2$, Zhuoran Wang$^2$, Baoxun Wang$^2$}
\address{$^1$Beijing University of Posts and Telecommunications, Beijing, China\\
  $^2$Tricorn (Beijing) Technology Co., Ltd, Beijing, China}
\email{bestbzw@bupt.edu.cn, \{yubo, wubowen, wangzhuoran, wangbaoxun\}@trio.ai}

\date{}

\begin{document}
\maketitle
\begin{abstract}

Generating structured query language (SQL) from natural language is an emerging research topic.
This paper presents a new learning paradigm from indirect supervision of the answers to natural language questions, instead of SQL queries.
This paradigm facilitates the acquisition of training data due to the abundant resources of question-answer pairs for various domains in the Internet, and expels the difficult SQL annotation job.
An end-to-end neural model integrating with reinforcement learning is proposed to learn SQL generation policy within the answer-driven learning paradigm.
The model is evaluated on datasets of different domains, including movie and academic publication. 
Experimental results show that our model outperforms the baseline models.
\end{abstract}

\section{Introduction}
\label{sec:intro}

Nowadays, task oriented dialogue systems allow intuitive interaction through natural language, where natural language understanding~(NLU) is an essential part. Structured Query Language~(SQL) is a standard language for accessing knowledge bases or relational databases. Thus, SQL generation from text is crucial for many NLU applications.
However, SQL is very difficult for users without technical training, thus natural language interfaces to databases have been widely studied~\cite{DB:DBLP:journals/nle/AndroutsopoulosRT95, DBLP:conf/coling/PopescuAEKY04, NLI-DB:Li:2014:CIN:2735461.2735468}. 
Most of these work adopts one or more of the following techniques, rule based pattern matching, syntactic grammars based parse tree mapping, semantic grammars based constituent tree mapping.
Some work~\cite{SemanticParsing:Clarke:2010:DSP:1870568.1870571, SemanticParsing:DBLP:conf/acl/LiangJK11, SemanticParsing:DBLP:conf/acl/CaiY13, SemanticParsing:Zettlemoyer:2005:LMS:3020336.3020416, SemanticParsing:Zettlemoyer07onlinelearning, SemanticParsing:DBLP:conf/emnlp/ArtziLZ15,SemanticParsing:DBLP:conf/acl/YihCHG15} is proposed as a subtask of semantic parsing.
These techniques focus on grammar parsing for specific domains, which cannot be easily generalized to different databases or application domains.


Several work on SQL generation from natural language (NL) has been proposed recently. A SQL generation model Seq2SQL is proposed in \cite{Seq2SQL} based on pointer networks~\cite{PointerNetworks}, together with a WikiSQL corpus of natural language questions, SQL queries and tables from Wikipedia. 
Some work~\cite{SQLNet,DBLP:journals/corr/abs-1804-09769} follows Seq2SQL and proposes various approaches to improve the performance of WikiSQL task. 
\cite{AAAI18:DBLP:journals/corr/abs-1711-06061} proposes an SQL generation model integrated with SQL grammar.
These work needs model training on datasets containing NL questions and corresponding SQL queries. Such data is hard to collect since SQL annotation requires a full knowledge of SQL grammars and the relations between all database tables. Therefore, we propose to learn SQL parsers from \textbf{indirect supervision}, where each NL sentence is labeled with the answer instead of the SQL query. This learning paradigm facilitates data acquisition, since the training data can be easily acquired from Internet or non-expert users' annotation.

In this paper, we propose a reinforcement learning based SQL generator (SQLGen), learned from indirect supervision of natural language questions and corresponding answers. SQLGen takes COPYNET~\cite{DBLP:conf/acl/GuLLL16}, an encoder-decoder structure as the neural network component. The policy based reinforcement learning is used to guide the learning of SQL generation, and two types of rewards are proposed. The rewards reflect the extent of correctness of generated SQL, which is an integration of correctness in the manner of logic and query execution.
In order to provide more precise supervision, the rewards are designed to be vectors instead of scalars, where each element is assigned to a corresponding word in the generated SQL query.

The main contributions of this paper are as follows.
(1) We propose a novel learning paradigm for SQL generation without annotated SQL queries for the first time. 
(2) We design an end-to-end neural model based on COPYNET with policy-based reinforcement learning for the answer-driven learning paradigm.
(3) We design a compound point-wise reward assignment mechanism for SQL generation policy learning.

\section{Related Work}
\label{sec:relatedwork}

Semantic parsing has attracted researchers' attention recent years, which refers to the problem of converting a natural language sentence to a formal meaning representation~\cite{SemanticParsing:Clarke:2010:DSP:1870568.1870571, SemanticParsing:DBLP:conf/acl/LiangJK11, SemanticParsing:DBLP:conf/acl/CaiY13}. Some research work focused on learning semantic parsers that generate logics executable on knowledge bases~\cite{SemanticParsing:Zelle:1996:LPD:1864519.1864543, SemanticParsing:Zettlemoyer:2005:LMS:3020336.3020416, SemanticParsing:Zettlemoyer07onlinelearning, SemanticParsing:DBLP:conf/emnlp/ArtziLZ15}. Recently, there has been some work attempting to learn parsers utilizing the results of query execution as indirect supervision~\cite{SemanticParsing:DBLP:journals/tacl/ReddyLS14, SemanticParsing:DBLP:conf/acl/YihCHG15, SemanticParsing:DBLP:conf/acl/PasupatL15, SemanticParsing:DBLP:conf/acl/GuuPLL17}. However, the grammar structure of SQL is much more complicated than the logical forms in semantic parsing~\cite{AAAI18:DBLP:journals/corr/abs-1711-06061}, and it is non-trivial to adapt the semantic parsing techniques to SQL generation domain. 


Although translating natural language into SQL queries has been extensively studied~\cite{DB:DBLP:journals/coling/WarrenP82, DB:DBLP:journals/nle/AndroutsopoulosRT95, DBLP:conf/coling/PopescuAEKY04, DBLP:conf/coling/GiordaniM12}, most work focuses on grammar parsing or interactive interface building which heavily relies on the grammar, and the proposed methods are difficult to be generalized to new databases. A neural system based on Seq2Seq model~\cite{Seq2Seq} is proposed~\cite{DBLP:conf/acl/IyerKCKZ17} to translate natural language to SQL queries with user feedbacks, which requires gathering user feedbacks to improve accuracy or adapt to new domains.
There has also been some work on answering natural language questions based on knowledge bases~\cite{QueryTables:DBLP:journals/debu/LuLK16, QueryTables:DBLP:journals/corr/MouLLJ16}.

The most relevant work includes the following. Seq2SQL~\cite{Seq2SQL} proposes a neural architecture based on pointer networks~\cite{PointerNetworks} to generate SQL queries with reinforcement learning. Seq2SQL also proposes a WikiSQL corpus of natural language questions, SQL queries and tables from Wikipedia source. SQLNet~\cite{SQLNet} follows the work of Seq2SQL and proposes a sequence-to-set-based approach without reinforcement learning, which improves the performance of WikiSQL task. 
TYPESQL~\cite{DBLP:journals/corr/abs-1804-09769} employs a slot filling model to predict the attribute values in SQL.
All methods split a SQL query into several parts, and predict each part using a different neural module. Furthermore, WikiSQL task only considers generating SQL queries with respect to one table. 
\cite{AAAI18:DBLP:journals/corr/abs-1711-06061} proposes an encoder-decoder framework integrated with SQL grammatical structures for SQL generation. It requires preprocessing of annotating the potential attribute values in natural language questions. Compared to the three methods, our approach has the following differences. (1) Our approach learns SQL queries with respect to multiple tables from indirect supervision of natural language question and answer pairs, instead of question and SQL pairs. (2) Our approach adopts an end-to-end learning framework without segmenting SQL queries and learning separately. 

Our work is also related to the work on attentional Seq2Seq models, which show promising performances on neural machine translation~\cite{AttentionalNMT:DBLP:journals/corr/BahdanauCB14, AttentionalNMT:DBLP:conf/acl/TuLLLL16}, dialog generation~\cite{Dialog:DBLP:conf/aaai/SerbanSLCPCB17, Dialog:DBLP:conf/acl/ShangLL15}, question answering~\cite{QA:DBLP:conf/acl/ChenFWB17, QA:DBLP:journals/corr/XiongZS16}, etc. Our work adopts the framework of COPYNET~\cite{DBLP:conf/acl/GuLLL16}, which incorporates the copying mechanism into the attentional encoder-decoder model. The intuition is that the words from the source sequence may appear in the target sequence, which is true for SQL generation.

\section{Task Description}
\label{sec:task}
The SQL generation task from natural language questions is described as follows. The input is the natural language question querying the database. The output is a SQL query, the meaning of which should be equivalent to that of the input question. 

\begin{figure}[t]
  \centering
  \setlength{\abovecaptionskip}{-5pt}
  \includegraphics[width=\linewidth]{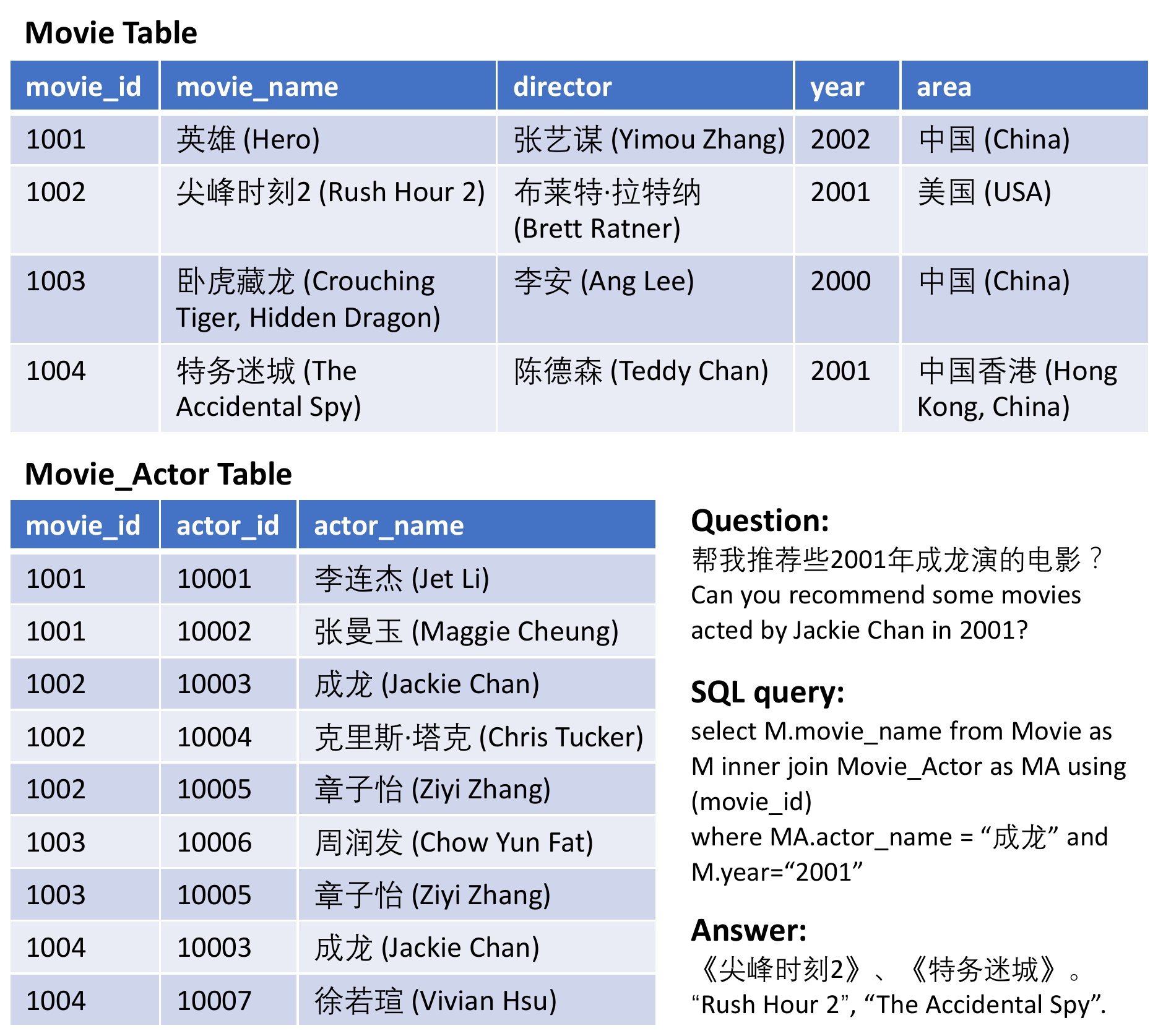}
  \caption{An example of SQL generation task. The two tables are sampled from a movie database. The question queries the movies acted by Jackie Chan in 2001, and the correct SQL query is shown. The information in the brackets of both tables are translations of the Chinese words.}
  \label{fig:example}
\end{figure}

We show an example in Figure~\ref{fig:example}. 
The ``Movie'' table contains the information of ``\emph{name}'', ``\emph{genre}'', ``\emph{director}'', ``\emph{year}'', ``\emph{vote}'' and ``\emph{language}'' of each movie, with ``\emph{name}'' as the primary key. 
The input question queries the names of movies in 2001 that are acted by Jackie Chan, and the output SQL query is shown in the figure where the table join operation is needed.

In order to make the problem more tractable, we make a similar assumption to WikiSQL, i.e., any non-SQL token in the generated SQL query should be a substring of the natural language question. Here the \textbf{SQL tokens} refer to all the SQL keywords (e.g. ``select'', ``from'', ``where'', etc.) and the names (including aliases) of tables and columns. For the example in Figure~\ref{fig:example}, the non-SQL tokens in the SQL query are ``Jackie Chan'' and ``2001'', which should appear in the question. This assumption also facilitates the utilization of COPYNET model, which learns to extract useful keywords from the questions.

Compared to WikiSQL task, our task has the following differences. (1) Our task learns from indirect supervision of the answers to natural language questions instead of SQL queries. (2) Our task considers generating a SQL query with respect to multiple tables, while WikiSQL considers only one table. 



\section{Approach}
\label{sec:approach}

In this section, we introduce our SQL generator SQLGen (shown in Figure~\ref{fig:model}), where an encoder-decoder based architecture COPYNET is employed for SQL generation. We also design a reward assignment mechanism based on the generated SQL queries and the answers. Thus, the generation policy can be supervised by reinforcement learning using the designed reward mechanism.

\subsection{Copying Mechanism for SQL Generation}
\label{subsec:copynet}

\begin{figure*}[th]
  \centering
  \setlength{\abovecaptionskip}{-2pt}
  \includegraphics[width=\linewidth]{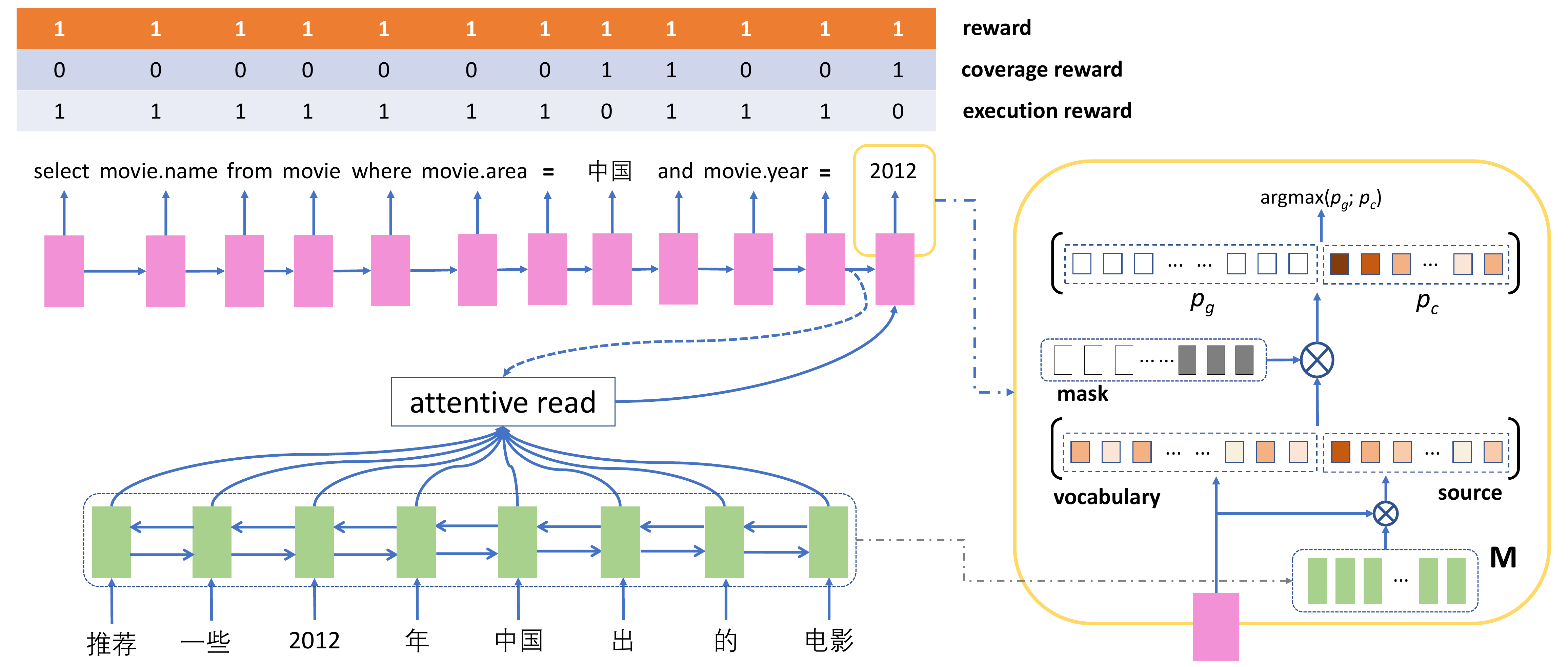}
  \caption{The overview of our SQL generator SQLGen. An example of SQL generation process is shown. The input natural language question asks to ``recommend some movies that are produced in China in 2012''. The SQL query is generated on the basis of a COPYNET structure, and the point-wise reward is computed for learning the generation policy by reinforcement learning.}
  \label{fig:model}
\end{figure*}

An encoder-decoder based framework COPYNET is employed, which incorporates the copying mechanism while decoding. 
As shown in Figure \ref{fig:model}, the input sequence of the natural language question is transformed by the encoder~(e.g. bidirectional RNN) into a representation $M$, and the decoder generates the output SQL query by predicting words based on a mixed probabilistic model of two modes, the generate-mode and the copy-mode. While decoding, COPYNET has not only an attentive read to $M$, but also a selective read to $M$, which renders the word generation from the designated vocabulary and the source sequence. 

\textbf{Vocabulary.} The vocabulary in SQL generation domain consists of two parts since the generated SQL query should contain both SQL tokens~(as defined in Section \ref{sec:task}) and non-SQL tokens~(the attribute values appeared in the source sequence). 

The first portion of the vocabulary is denoted by $V_{SQL}$, which contains the SQL keywords, operators and database symbols. 
\begin{itemize}
\setlength{\itemsep}{-1pt}
\item The SQL keyword set $V_{key}$ contains all the SQL keywords, such as ``select'', ``where''. 
\item The comparator set $V_{cmp}$ contains all the comparative operators, e.g. ``$=$'', ``$>$'', etc.
\item The database symbol set $V_{db}$ contains all the names of database tables and columns.
\end{itemize}

Here we further introduce the constituents of the database symbol set $V_{db}$. Let the \textbf{table set} of the database be $T=\{T_1, T_2, \cdots, T_t\}$, where $T_i$ is the name of the $i$-th table. Let the \textbf{column set} with respect to table $T_i$ be $Col_i=\{col_{ij}\}$, where $col_{ij}$ is the name of the $j$-th column in table $T_i$. The elements in both $T$ and $\{Col_i\}$ are database symbols. In order to reduce the exploring space of reinforcement learning, we further clarify the column set by introducing the \textbf{attribute set} $Attr_i=\{T_i.col_{ij}|j=1,\cdots,|Col_i|\}$. Take the example in Figure~\ref{fig:example}, the attribute set for the ``Movie'' table is $\{$``movie.movie\_id'', ``movie.movie\_name'', ``movie.director'', $\cdots\}$. The database symbol set $V_{db}$ covers the table set $T$ and the attribute sets $\{Attr_i\}$.

Thus, $V_{SQL}=V_{key}\cup V_{cmp}\cup V_{db}$. 

The second portion of the vocabulary is denoted by $V_{NL}$, which covers all the unique words that appear in the natural language questions. Therefore, the whole vocabulary $\mathcal{V}$ is $V_{SQL}\cup V_{NL}$.

\textbf{Encoder.} Let $\mathcal{X} = \{x_1,\cdots,x_n\}$ be the input sequence. As shown in Figure \ref{fig:model}, the input sequence $\mathcal{X}$ (``Recommend some movies that are produced in China in 2012'') is converted into a representation $M=\{h_1, \cdots, h_n\}$ by an RNN encoder as follows. Note that a bidirectional GRU~\cite{GRU} is used in this work. 
{\setlength\abovedisplayskip{6pt}
\setlength\belowdisplayskip{6pt}
\begin{equation}
h_t=\text{BiGRU}(h_{t-1}, x_t)
\end{equation}}
The representation $M$ will be accessed by the decoder during the process of SQL generation.

\textbf{Decoder.} A GRU layer is used as the decoder to predict the target sequence. Let the decoder states be $\{s_t\}$ and the generated words be $\{y_t\}$. We apply a standard attention mechanism on $M$ and obtain a context vector sequence $C=\{c_t\}$.


Given the decoder state $s_t$, context vector $c_t$ and $M$, the probability of generating a word $y_t$ is computed as follows.
{\setlength\abovedisplayskip{5pt}
\setlength\belowdisplayskip{5pt}
\begin{align}
\nonumber p(y_t|s_t, y_{t-1}, c_t, M)& = p(y_t,\mathbf{g}|s_t, y_{t-1}, c_t, M) + \\
 &p(y_t,\mathbf{c}|s_t, y_{t-1}, c_t, M)
\end{align}}
where $\mathbf{g}$ stands for the generate-mode, and $\mathbf{c}$ for the copy-mode. The probabilities for the two modes are computed as follows.
{\setlength\abovedisplayskip{4pt}
\setlength\belowdisplayskip{4pt}
\begin{equation*}
p(y_t , \mathbf{g} |\cdot) =
\begin{cases}
\frac{1}{Z}e^{\psi_g(y_t)}\quad\qquad\qquad &\text{if $y_t\in{V_{SQL}}$}\\
0 &\text{otherwise}
\end{cases}
\end{equation*}}
{\setlength\abovedisplayskip{4pt}
\setlength\belowdisplayskip{4pt}
\begin{equation*}
p(y_t , \mathbf{c} |\cdot) =
\begin{cases}
\frac{1}{Z}\sum_{j:x_j=y_t}{e^{\psi_c(x_j)}} &\text{if $y_t\in \mathcal{X}$}\\
0 &\text{otherwise}
\end{cases}
\end{equation*}}
where $Z$ is the normalization term shared by the generate-mode and copy-mode as follows. 
{\setlength\abovedisplayskip{5pt}
\setlength\belowdisplayskip{4pt}
\begin{equation}
Z=\sum_{v\in V_{SQL}}e^{\psi_g(v)} + \sum_{x\in \mathcal{X}}e^{\psi_c(x)}
\end{equation}}
$\psi_g(\cdot)$ and $\psi_c(\cdot)$ are scoring functions as follows, for generate-mode and copy-mode, respectively.
{\setlength\abovedisplayskip{5pt}
\setlength\belowdisplayskip{0pt}
\begin{equation}
\psi_g (y_t = v_i) = \mathbf{v_i^\mathrm{T}} W_o s_t \qquad v_i \in \mathcal{V}_{SQL}
\end{equation}}
{\setlength\abovedisplayskip{0pt}
\setlength\belowdisplayskip{5pt}
\begin{equation}
\psi_c (y_t = x_j ) = \sigma(h_j^\mathrm{T} W_c)s_t \qquad x_j \in \mathcal{X}
\end{equation}}
where $W_o$ and $W_c$ are learnable parameters, and $\mathbf{v_i}$ is the one-hot indicator vector for $v_i$.

Note that a specific state update mechanism is introduced in COPYNET, which can be eliminated if Chinese word segmentation or English chunking is done during preprocessing, or reserved otherwise. The state update mechanism helps to copy a consecutive sub-sequence in the source text, while an attribute value to be copied should be words in a single chunk after preprocessing in our task. 

\textbf{Mask.} We rely on reinforcement learning to learn the generation policy since there is no correct SQL queries as direct supervision. However, the exploration space is enormous due to the complexity of the natural language and SQL logic. To solve this problem, we introduce a masking mechanism to reduce the exploration space. When the decoder is predicting the next target word, a mask vector $\mathbf{m}=(m_1, \cdots, m_k)$ is introduced to indicate whether a word is legal for generation given the previous word(s), as illustrated in Figure \ref{fig:model}. The dimension $k$ of $\mathbf{m}$ is $|V_{SQL}|+|\mathcal{X}|$, and $m_t=1$ if word $v_t$ is legal, $m_t=0$ otherwise. 

The mask mechanism can be easily implemented based on SQL grammar. For example, if the previous generated word is the SQL keyword ``from'', the current word should be the name of a certain table, thus the other words are illegal. Therefore, the mask mechanism helps to generate grammatically correct SQL queries.

\subsection{Reinforcement Learning with Compound Reward}
\label{subsec:rl}

We apply reinforcement learning to learn a SQL generation policy under the indirect supervision of answers. Unlike \cite{Seq2SQL}'s work, which assigns a scalar reward to a generated SQL query, we design a compound point-wise reward that acts on each token of the generated SQL query. This mechanism helps to guide the learning of SQL generation policy more precisely.

\begin{figure}[t]
  \centering
  \setlength{\abovecaptionskip}{-5pt}
  \includegraphics[width=\linewidth]{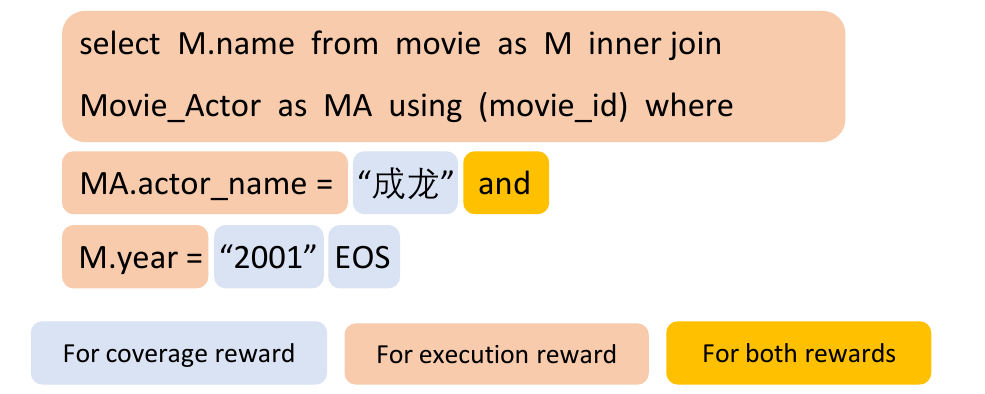}
  \caption{An illustration of two types of rewards, which act on different parts of the SQL query.}
  \label{fig:rewards}
\end{figure}

The point-wise reward mechanism is composed of two types of rewards, the coverage reward and the execution reward, which are acted on different portions of SQL queries. As illustrated in Figure~\ref{fig:rewards}, the coverage reward is acted on the words of attribute values in the where-conditions, the operators~(``and'', ``or'') connecting where-conditions, and the token for end-of-sentence (EOS), while the execution reward is acted on all the other words and the operators as in coverage reward.

\textbf{Coverage reward.} The coverage reward aims to guide the learning of word selection from the source text, and the procedure of coverage computation is shown in Algorithm 1. In order to better supervise the copy-mode learning of COPYNET, a vocabulary of attribute values is extracted from the database, which covers the possible values of queried attributes. Thus, the attribute values in the source text can be obtained based on this attribute-value vocabulary. The correct copied words in the generated SQL query are assigned positive rewards of $1$, while the incorrect words and the duplicate correct words are assigned negative rewards of $-1$. 

Similarly, the correct operators in the generated SQL query are assigned equally positive rewards, while incorrect operators are assigned non-positive rewards. Since there is no direct supervision of the correct SQL query, it is impossible to know whether a generated operator is semantically correct. What we know is the number $K$ of attributes in the correct SQL based on the source text and the attribute-value vocabulary. Hence, a \textbf{correct operator} here refers to the first $K$ operators in the generated SQL, while an \textbf{incorrect operator} refers to the other redundant operators. The first incorrect operator is assigned a negative reward of $-1$, leaving the others no penalty in case that the operators are excessively penalized.

For the EOS token, we reward EOS in the SQL queries with the correct number of attributes and penalize EOS in those with insufficient number of attributes, leaving EOS in other cases no penalty.

\begin{algorithm}
  \label{alg:coverage}
  \caption{Coverage reward computation}
  \begin{algorithmic}[1]
    \REQUIRE SQL query $Q$, Source text $S$
    \ENSURE Coverage reward $R_c$
    \STATE $U \gets$\text{ the set of attribute values in source text}
    \STATE $V \gets \emptyset$
    \FOR{$w$ in copied words in $Q$}
      \IF{$w \in U$ \AND $w \notin V$}
        \STATE $R_c(w)\gets 1, V\gets V \cup \{w\}$
      \ELSE
        \STATE Set $R_c(w)$ to $-1$
      \ENDIF
    \ENDFOR
    \FOR{$l$-th operator $op_l$ in $Q$}
      \STATE Set $R_c(op_l)=\begin{cases}1/(|U|-1) &\text{if } l<|U|\\-1 &\text{if } l=|U|\\0 &\text{if } l>|U|\end{cases}$
    \ENDFOR
    \STATE $N_{op}\gets$ the number of operators in $Q$
    \STATE Set $R_c(EOS)=\begin{cases}-1 &\text{if } N_{op}<|U|-1\\1 &\text{if } N_{op}=|U|-1\\0 &\text{if } N_{op}>|U|-1\end{cases}$
  \end{algorithmic}
\end{algorithm}

\textbf{Execution reward.} The execution reward aims to guide the learning of SQL representation for natural language logics. The procedure of execution reward computation is shown in Algorithm 2. The execution rewards act on three types of SQL segments, the text segment $b$ from ``select'' to ``where'', the condition-clauses $C'$ without attribute values, operators $O$ connecting condition-clauses. For the example in Figure \ref{fig:rewards}, $b$ is ``select $\cdots$where'', $C'$ is \{``MA.actor\_name='',``M.year=''\}, $O$ is \{``and''\}. The words in these SQL segments constitute the targeted word set for the execution reward.

The generated SQL query $Q$ is executed. If the query result is equal to the answer, it is believed that $Q$ is correctly generated and the rewards for the targeted words of $Q$ is set to $1$. Otherwise, the rewards of words in $b$ are set to $-1$, while those of words in $O$ are set to $0$. For $c'_i$ in $C'$, the SQL query with corresponding single condition is executed. If the result and answer set $A$ have common elements, the rewards are set to $1$ since the attribute-value pair in the condition should be correct, $-1$ otherwise. In this way, execution reward guides the reinforcement learning model by assigning higher rewards to correct SQL queries. 

Note that we assume the form of the condition clause to be ``attribute=value'', which restricts the comparator to ``=''. The reasons are two fold. First, the value types are mostly strings in our movie domain, thus equality is the most common comparator, while there is rare data with other comparators. Second, considering all comparators significantly raises the learning complexity, which we hope to study in our future work.

For a SQL query $Q$, the whole point-wise reward $R$ is a combination of the coverage reward $R_c$ and the execution reward $R_e$, which act on word set $V_c(Q)$ and $V_e(Q)$, respectively. As described above, $V_c(Q)\cap V_e(Q)=O(Q)$, which is the set of the operators $O(Q)$ connecting condition clauses. The whole reward $R(w)$ for each $w$ in $Q$ is computed as follows.
{\setlength\abovedisplayskip{4pt}
\setlength\belowdisplayskip{4pt}
\begin{equation*}
R(w) =
\begin{cases}
R_c(w) &\text{if } w\in V_c\setminus O\\
R_e(w) &\text{if } w\in V_e\setminus O\\
\min\{R_c(w), R_e(w)\} &\text{if $w\in O$}
\end{cases}
\end{equation*}}

\begin{algorithm}
  \label{alg:execution}
  \caption{Execution reward computation}
  \begin{algorithmic}[1]
    \REQUIRE SQL query $Q$, Answer set $A$
    \ENSURE Execution reward $R_e$
    \STATE Segment $Q$ by ``where'' and operators
    \STATE $b \gets$\text{ text from ``select'' to ``where''}
    \STATE \textit{\# condition clause form: ``attribute=value''}
    \STATE $C \gets$\text{ the set of condition-clauses}
    \STATE $C' \gets \{c'_i\text{=substring ``attribute='' of } c|c\in C\}$
    \STATE $O \gets \{\text{operators connecting clauses in C}\}$
    \STATE Execute SQL query $Q$ and get result $Res$
    \IF{$A=Res$}
      \STATE Set $R_e$ for words in $\{b\}\cup C'\cup O$ to $1$
      \RETURN
    \ENDIF
    \STATE Set $R_e$ for words in $b$ to $-1$
    \STATE Set $R_e$ for words in $O$ to $0$
    \FOR{$c_i$ in $C$}
      \STATE Concatenate $b$ with $c_i$, get SQL query $Q_i$
      \STATE Execute SQL query $Q_i$, get result $Res_i$
      \IF{$Res_i\cap A \neq \emptyset$}
        \STATE Set $R_e$ for words in $c'_i$ to $1$
      \ELSE
        \STATE Set $R_e$ for words in $c'_i$ to $-1$
      \ENDIF
    \ENDFOR
  \end{algorithmic}
\end{algorithm}

\textbf{Learning.} We define the accumulative reward of SQL query $Q=[q_1,\cdots,q_T]$ to be $\tilde{R}(Q)=\sum_{i}R(q_i)$. The loss function is the negative expected accumulative reward over possible SQL queries, i.e., $L=-\mathbb{E}(\tilde{R}(Q))$. We have the following equality as shown in \cite{DBLP:conf/nips/SchulmanHWA15}.
{\setlength\abovedisplayskip{4pt}
\setlength\belowdisplayskip{4pt}
\begin{equation*}
\begin{aligned}
\nabla_\Theta(\mathbb{E}_y(R(y))) 
= \mathbb{E}_y(R(y)\cdot\nabla_\Theta\log p(y;\Theta))
\end{aligned}
\end{equation*}}
Thus, the policy gradient of the loss function $L$ can be derived as follows. We approximate the expected gradient with a single Monte-Carlo sample $Q$ in the last step of the derivation.
{\setlength\abovedisplayskip{4pt}
\setlength\belowdisplayskip{-4pt}
\begin{equation*}
\begin{aligned}
\nabla_\Theta(L) & = -\nabla_\Theta\mathbb{E}_{Q\sim p(Q)}(\sum_{i}R(q_i))\\
& = -\sum_{i}\nabla_\Theta\mathbb{E}_{Q\sim p(Q)}(R(q_i)) \\
& = -\sum_{i}\mathbb{E}_{Q\sim p(Q)}(R(q_i) \nabla_\Theta\log p_Q(q_i;\Theta)) \\
& = -\mathbb{E}_{Q\sim p(Q)}(\nabla_\Theta\sum_{i}(R(q_i) \log p_Q(q_i;\Theta))) \\
& \approx -\nabla_\Theta\sum_{i}(R(q_i) \log p_Q(q_i;\Theta))
\end{aligned}
\end{equation*}}

\section{Experiments}
\label{sec:experiments}

\textbf{Data.} We collect the datasets for evaluation, including a Chinese dataset in movie domain, two English datasets in the domains of academic publication and movie. The datasets consist of natural language questions, corresponding answers and database tables. For the comparison with direct supervised learning methods, we ask volunteers to label the questions with SQL queries. 

\emph{Movie-Chinese dataset}. The questions and answers are collected from a Chinese QA community (Baidu Zhidao), and the database is constructed using data collected from a Chinese movie community (Douban). 
There are 3 tables in the database, containing information of actors, directors, types, areas and languages of movies.
We preprocess the data to eliminate the illegal data, such as confusing questions, incorrect answers. 


The proportion of questions involving multiple tables is 78\%, while that involving multiple conditions is 43\%.
The questions involving multiple tables have a high proportion because most SQL queries contain at least the ``movie'' table, since users tend to query their interested movie names.

Different from \textit{Movie-Chinese}, the other two datasets are synthetic, where the databases are constructed by data collection from Internet and question-answer pairs are generated by templates.

\emph{Academic dataset}. The database is constructed using the data from \cite{Roy2013The}, where we select 3 tables for our task, containing records of papers, researchers and conferences. 

\emph{Movie dataset}. The database is constructed using an open-source dataset\footnote{https://github.com/sundeepblue/movie\_rating\_prediction} of IMDB. The dataset contains the same attributes as \textit{Movie-Chinese}.

Each dataset contains around 10,000 question-answer pairs, and is randomly partitioned into training set, validation set and test set with the proportion of $80\%$, $10\%$, $10\%$, respectively. \textbf{The datasets can be referred to the supplementary materials in our submission.}

\textbf{Baselines.} (1) Seq2Seq-RL is an attentional Seq2Seq model with reinforcement learning using our point-wise rewards. (2) CopyNet-Seq2SQL is a COPYNET model with reinforcement learning using rewards in Seq2SQL~\cite{Seq2SQL}. (3) CopyNet-SL is a COPYNET model supervised by the annotated SQL queries.

We also study the performance of SQLGen with pretraining by the annotated SQL queries, which we denote by SQLGen-Pretrain.

\textbf{Evaluation.} Two evaluation metrics are used, accuracy and redundancy. 
\textit{Accuracy} refers to the ratio of correct SQL queries, where a query is correct if it executes to the correct result. \textit{Redundancy} refers to the ratio of redundant SQL queries, where a query is redundant if it joins the tables that are in none of the conditions.

\textbf{Settings.} The hidden unit size of the encoder and decoder is 32 and 64, respectively. The embedding size is set to 50 due to a small vocabulary size. The models are trained at most 100 epochs with early stopping, using the Adam optimizer. While decoding, we either randomly sample a word from the distribution with probability $\epsilon$, or pick the highest-scoring word with the probability $1-\epsilon$, rendering reinforcement learning more exploration opportunities. We set $\epsilon=0.3$ in the experiments.

\begin{table*}[htbp]
  \label{tab:result1}
  \setlength{\abovecaptionskip}{3pt}
  \setlength{\belowcaptionskip}{-1pt}
  \centering
\begin{tabular}{c|cc|cc|cc}  
   \hline
   \multirow{2}*{Models} &\multicolumn{2}{c|}{\textit{Movie-Chinese}}&\multicolumn{2}{c|}{\textit{Academic}}&\multicolumn{2}{c}{\textit{Movie}}\\
   \cline{2-7}
               & Accuracy &Redundancy  & Accuracy &Redundancy& Accuracy &Redundancy\\
\hline
Seq2Seq-RL        & 8.1 & 24.7 &0.0 &  -   &0.0  & - \\
CopyNet-Seq2SQL        & 17.0& 100.0&2.5 &29.4  &0.0  & - \\
CopyNet-SL   & 56.9& 0.2  &62.6&0.0   &51.9 &0.0\\

\hline
\textbf{SQLGen}& 59.8&68.6  &64.6&70.2  &70.0 &76.0\\
SQLGen-Pretrain& 80.4&75.4  &67.8&99.4 &73.6  &97.4\\
\hline
\end{tabular}
\caption{The accuracy and redundancy of SQLGen and the baselines on three datasets.}
\end{table*}

\subsection{Main Results}

Table 1 shows the accuracy and redundancy results of SQLGen and the baselines on the three datasets.
The first two baselines, Seq2Seq-RL and CopyNet-Seq2SQL, have very low accuracy. This result shows the difficulty of the proposed learning paradigm with indirect supervision. We also try the Seq2Seq model with the typical Seq2SQL reward, which hardly learns anything and have an accuracy of $0$, thus we do not take it as a baseline model.
CopyNet-SL has better performances than the other two baselines since it learns from the direct supervision of correct SQL queries. 
SQLGen has higher accuracy than CopyNet-SL. A probable reason is that CopyNet-SL learns from supervised SQL queries but penalizes correct SQL queries with different orders of table joins or conditions.

SQLGen-Pretrain have higher accuracy than SQLGen by 1\%-34\% for different datasets. This demonstrates that supervised pretraining helps improve the subsequent policy learning but needs the manual annotation. Thus, a suitable method can be selected based on a tradeoff between performance and annotation cost in practical scenarios.

We study the redundancy of different models when the accuracy is higher than $0$. SQLGen has the redundancy of 68\%-75\% on different datasets. The reason is that the space for the combinations of table joins and conditions is enormous and it is very difficult for the indirect supervised learning. Thus, the exploration of reinforcement learning has a tendency of joining more potential tables, which has a higher probability of generating correct SQL queries. This tendency results in relatively high redundancy. Note that SQLGen does not generate SQL queries with duplicate tables or conditions using the mask proposed in Section \ref{subsec:copynet}.

CopyNet-SL has very low redundancy since the model learns from direct supervision of SQL queries, which has a low probability of joining redundant tables. SQLGen-Pretrain has higher redundancy than SQLGen. The reason is that the pretrained model has a tendency of joining more tables than a randomly initialized model, since the training data involving multiple tables takes a high proportion, as shown in the data description.

Table 2 shows the accuracy on \textit{Movie-Chinese} dataset in different cases, including SQL queries containing single and multiple conditions (tables).
For SQLGen and the baselines, the accuracy of SQL queries with single condition is higher than that with multiple conditions, because the natural language related to a single condition is easier to learn. 
SQLGen has much lower accuracy for SQL queries with single table than those with multiple tables. By observing the test cases, we find most of the incorrect SQL queries predict the wrong attributes in the condition clauses. As shown in Figure \ref{fig:cases}, the attribute is difficult to learn since the patterns querying different attributes could be similar due to the characteristics of Chinese language. In \textit{Movie-Chinese} domain, such patterns mostly occur in the cases where single table is involved.

\begin{table}[t]
  \label{tab:result2}
  \setlength{\abovecaptionskip}{3pt}
  \setlength{\belowcaptionskip}{0pt}
  \centering
\begin{tabular}{p{3.05cm}p{0.7cm}<{\centering}p{0.7cm}<{\centering}p{0.7cm}<{\centering}p{0.7cm}<{\centering}} 
   \hline
     Models           & $\text{Acc}_\text{sc}$ &$\text{Acc}_\text{mc}$& $\text{Acc}_\text{st}$ &$\text{Acc}_\text{mt}$\\ 
     \hline
     Seq2Seq-RL         & 13.6       &0.7     &1.1         &9.7     \\
     CopyNet-Seq2SQL    & 29.7       & 0.0     &0.0         &21.0     \\
     CopyNet-SL         & 65.6       & 45.2    &90.4        &49.2    \\
     \hline
     \textbf{SQLGen}    & 61.8       & 57.1    &34.2        &65.7    \\
     SQLGen-Pretrain          & 91.1        &57.6    &90.9        &73.6    \\
     \hline
\end{tabular}
\caption{The accuracy analysis on \textit{Movie-Chinese} dataset. $\text{Acc}_\text{sc}$($\text{Acc}_\text{mc}$) is the accuracy of SQL queries with single (multiple) condition(s). $\text{Acc}_\text{st}$($\text{Acc}_\text{st}$) is that with single (multiple) table(s).}
\end{table}

\begin{figure}[t]
  \centering
  \setlength{\abovecaptionskip}{-15pt}
  \setlength{\belowcaptionskip}{-2pt}
  \includegraphics[width=\linewidth]{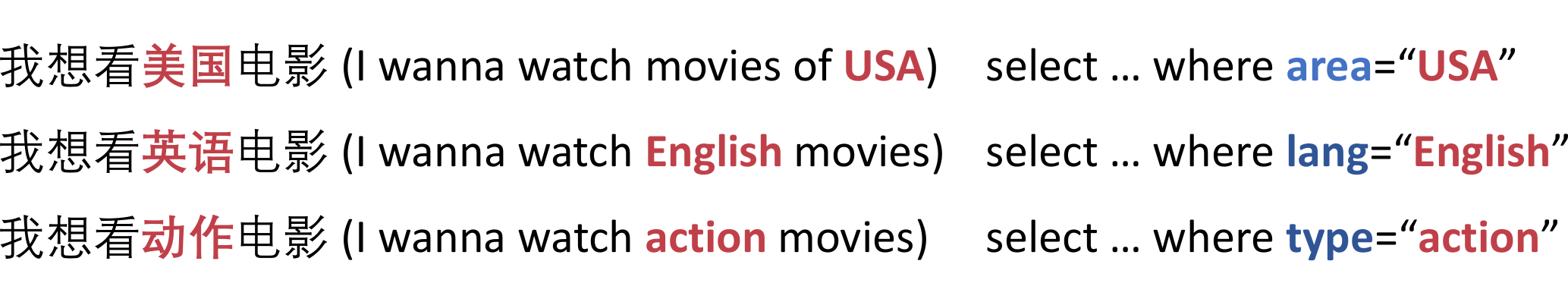}
  \caption{An illustration of similar patterns.}
  \label{fig:cases}
\end{figure}

Compared to CopyNet-SL, SQLGen shows higher accuracy on SQL queries with multiple conditions (tables) but lower accuracy for single condition (table), because CopyNet-SL penalizes correct SQLs with multiple conditions (tables) with different orders from the training data. 
SQLGen-Pretrain outperforms SQLGen by better learning attributes of values in natural language, which helps to improve the accuracy for single condition and table.

\section{Conclusion and Future Work}
\label{sec:conclusion}

We propose a SQL generation learning paradigm from indirect supervision of question-answer pairs in this paper. A COPYNET-based neural model integrating policy-based reinforcement learning is proposed, where a compound reward mechanism is designed to precisely learn the generation policy. Experimental results show that our model has higher accuracy than baselines on various datasets. 

In the future work, we would like to design models that can generate more complex SQL queries, e.g. queries with more operators and comparators in the condition clauses.




\bibliography{acl2018}
\bibliographystyle{acl_natbib}


\end{document}